\documentclass[final]{cvpr}

\usepackage{times}
\usepackage{epsfig}
\usepackage{graphicx}
\usepackage{amsmath}
\usepackage{amssymb}

\usepackage{pifont}
\usepackage{multirow}
\usepackage{algorithm}
\usepackage{algpseudocode}

\usepackage[pagebackref=true,breaklinks=true,colorlinks,bookmarks=false]{hyperref}

\newcommand{\LL}{\mathcal{L}}

\def\cvprPaperID{562} 
\def\confYear{CVPR 2021}
\begin{document}

\title{S$^2$-BNN: Bridging the Gap Between Self-Supervised Real and 1-bit Neural Networks via Guided Distribution Calibration}

 \author{Zhiqiang Shen$^1$,~Zechun Liu$^{1,2}$,~Jie Qin$^{3}$,~Lei Huang$^{3}$,~Kwang-Ting Cheng$^{2}$,~Marios Savvides$^1$ \\
	$^1$Carnegie Mellon University~~$^2$Hong Kong University of Science and Technology\\$^3$Inception Institute of Artificial Intelligence\\
	\tt\small{\{zhiqians,zechunl,marioss\}@andrew.cmu.edu}\\\tt\small{\{qinjiebuaa,huanglei36060520\}@gmail.com} \tt\small{timcheng@ust.hk}
}

\maketitle


\begin{abstract}
Previous studies dominantly target at self-supervised learning on real-valued networks and have achieved many promising results. However, on the more challenging binary neural networks (BNNs), this task has not yet been fully explored in the community. In this paper, we focus on this more difficult scenario: learning networks where both weights and activations are binary, meanwhile, without any human annotated labels. We observe that the commonly used contrastive objective is not satisfying on BNNs for competitive accuracy, since the backbone network contains relatively limited capacity and representation ability. Hence instead of directly applying existing self-supervised methods, which cause a severe decline in performance, we present a novel guided learning paradigm from real-valued to distill binary networks on the final prediction distribution, to minimize the loss and obtain desirable accuracy. Our proposed method can boost the simple contrastive learning baseline by an absolute gain of {\bf \em 5.5$\sim$15\%} on BNNs. We further reveal that it is difficult for BNNs to recover the similar predictive distributions as real-valued models when training without labels. Thus, how to calibrate them is key to address the degradation in performance. Extensive experiments are conducted on the large-scale ImageNet and downstream datasets. Our method achieves substantial improvement over the simple contrastive learning baseline, and is even comparable to many mainstream supervised BNN methods. Code is available at \url{https://github.com/szq0214/S2-BNN}.
\end{abstract}

\vspace{-0.3in}
\section{Introduction}

\begin{figure}[t]
  \centering
  \includegraphics[width=0.49\textwidth]{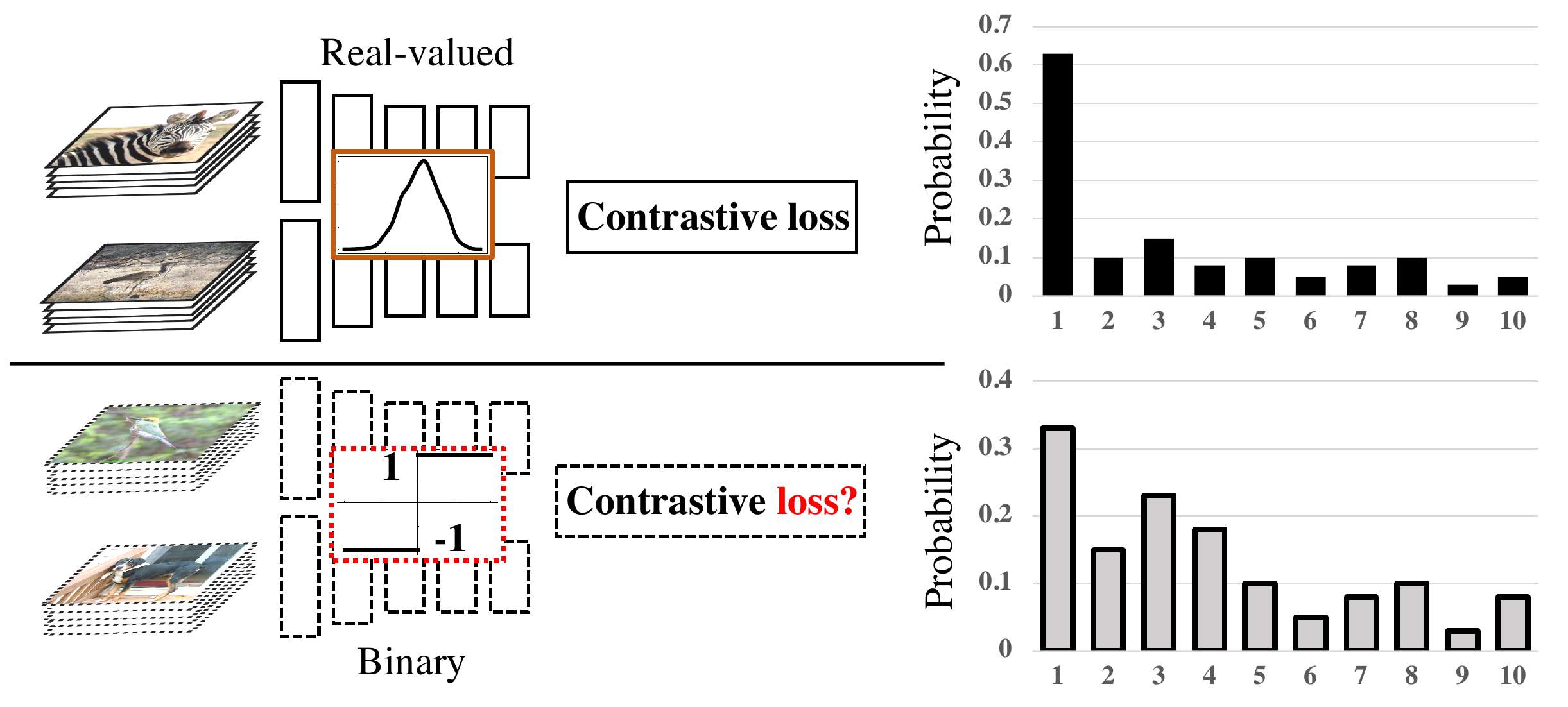}
  \vspace{-0.17in}
  \caption{Illustration of the motivation for self-supervised BNNs. As the representation capability of BNNs is relatively limited, the output prediction is less confident than that of real-valued networks. Hence, we argue that the conventional self-supervised methods in real-valued networks may not be optimal for BNNs.}
  \label{fig:motivation}
  \vspace{-0.12in}
\end{figure}

The recent advances and breakthroughs in 1-bit convolutional neural networks (1-bit CNNs), also known as binary neural networks~\cite{courbariaux2016binarized,rastegari2016xnor} mainly lie in supervised learning~\cite{liu2020reactnet,martinez2020training}. With the binary nature of BNNs, such networks have been recognized as one of the most efficient and promising deep compression techniques for deploying models in resource-limited devices. Generally, as introduced in~\cite{rastegari2016xnor}, BNNs can produce up to 32$\times$ compressed memory and 58$\times$ practical computational reduction on a CPU or mobile device. Considering the immense potential of being directly deployed in intelligent devices or low-power hardware, it is well worth further studying the behaviors of self-supervised BNNs ({\bf S$^2$-BNN}), i.e., BNNs without human-annotated labels, both to better understand the properties of BNNs in academia, as well as to extend the scope of their usage in industry and real-world applications.

\begin{table*}[t]
\centering
\caption{A brief overview of our improvement over a variety of different architectures for binarizing models. We choose XNOR Net~\cite{rastegari2016xnor}, Bi-Real Net~\cite{liu2018bi} and ReActNet~\cite{liu2020reactnet} as our backbone networks.} 
\label{tab:my-table_overview_1}
\resizebox{1.0\textwidth}{!}{
\begin{tabular}{lcccccc}
\hline
       & \multicolumn{2}{c}{XNOR (Re-impl.)~\cite{rastegari2016xnor}}                    & \multicolumn{2}{c}{Bi-Real Net~\cite{liu2018bi}}             & \multicolumn{2}{c}{ReActNet~\cite{liu2020reactnet}}                  \\
         & Top-1                & Top-5                & Top-1                & Top-5                & Top-1                  & Top-5                \\ \hline
Supervised BNN                 & 51.200                 &       73.200         & 56.400                 & 79.500                &    69.400            &     88.600           \\ \hline   
Contrastive Learning (MoCo V2) - Real-valued           &    --      &      --       & 50.296               & 75.206               & 60.776                 & 82.830               \\
Contrastive Learning (MoCo V2) - BNN (Baseline) &    23.880    &    44.690      & 42.816               & 67.712               & 46.922                & 70.712               \\ \hline
Contrastive Learning w/ Adam + lite aug. + progressive binarizing etc. (Ours) - BNN     &    --      &     --      &    --    &     --      & 52.452        &   76.080        \\ 
 + Guided Learning (Ours) - BNN     &     --        &       --       &      --      &      --       &    56.022  &  79.168   \\ 
Guided Learning Only (Ours) - BNN   &    \bf 36.996        & \bf 61.416     &  \bf 51.242  &  \bf 75.890     &  \bf 61.506 &  \bf 83.512 \\ \hline

\end{tabular}
}
\vspace{-0.15in}
\end{table*}

The goal of this paper is to study the mechanisms and properties of BNNs under the self-supervised learning scenario, then deliver practical guidelines on how to establish a strong self-supervised framework for them. To achieve this purpose, we start from exploring the widely used self-supervised contrastive learning in real-valued networks. Hence, our first question in this paper is: {\em Is the well-performing contrastive learning in real-valued networks still suitable for self-supervised BNNs?} Intuitively, binary networks are quite different from real-valued networks on both learning optimization and back-propagation of gradients since the weights and activations in BNNs are discrete, causing dissimilar predictions between the two different types of networks, as illustrated in Fig.~\ref{fig:motivation}. We answer this question by exploring the optimizers (SGD or the adaptive Adam optimizer), learning rate schedulers, data augmentation strategies, etc., and give optimal designs for self-supervised BNNs. These non-trivial studies enable us to build a base solution which brings about 5.5\% improvement over the na\"ive contrastive learning of the baseline.

Subsequently, we empirically observe that the real-valued networks always achieve much better performance than BNNs on self-supervised learning (the comparison will be given later). Many recent studies~\cite{liu2020reactnet,martinez2020training} have shown that BNNs demonstrate sufficient capability to achieve accuracy as high as the real-valued counterparts in supervised learning, but an appropriate learning strategy is required to unleash the potential of binary networks. Our second question is thus: {\em What are the essential causes for the performance gap between real-valued and binary neural networks in self-supervised learning?} It is natural to believe that if we can expose the causes behind the inferior results and also find a proper method for training self-supervised BNNs to mitigate the obliterated/poor accuracy, we can categorically obtain more competitive performance for self-supervised BNNs. Our discovery on this perspective is interesting: we observe that the distributions of predictions from BNNs and real-valued networks are significantly different but after using a frustratingly simple method through a {\em teacher-student} paradigm to calibrate the latent representation on BNNs, the performance of BNNs can be boosted substantially, with an extra $\sim$4\% improvement. 

Concretely, to address the issue of how to maximize the representation ability of self-supervised BNNs, we propose to add an additional self-supervised real-valued network branch to guide the target binary network learning. This is somewhat like knowledge distillation but the slight difference is that our teacher is a self-supervision learned network and the class for the final output is agnostic. We force the BNNs to mimic the final predictions of real-valued models after the projection MLP head and the softmax operation. In our framework, we introduce a strategy that enables the BNNs to mimic the distribution of a real-valued reference network smoothly. This procedure is called guided distillation in our method. Combining contrastive and guided learning is a spontaneous idea for tackling this problem, while intriguingly, we further observe that solely employing guided learning without contrastive loss can dramatically boost the performance of the target model by an additional 5.5\%. This is surprising since, intuitively, combining both of them seems a better choice. To shed further light on this observation, i.e., contrastive learning is not necessary for directly training self-supervised BNNs, we study the learning mechanism behind contrastive and guided/distilled techniques and derive the insights that contrastive and guided learning basically focus on different aspects of feature representations. Distillation forces BNNs to mimic the reference network's predictive probability, while contrastive learning tries to discover and learn the latent patterns from the data itself. This paper does not argue that learning the isolated patterns by contrastive learning is not good, but from our experiments, it shows that recovering knowledge from a well-learned real-valued network with extremely high accuracy is more effective and practical for self-supervised BNNs. An overview of our improvement over various architectures is shown in Table~\ref{tab:my-table_overview_1}.

To summarize, our contributions in this paper are:
\vspace{-0.06in}
\begin{itemize}
	\addtolength{\itemsep}{-0.1in}
	\item We are the first to study the problem of self-supervised binary neural networks. We provide many practical designs, including optimizer choice, learning rate scheduler, data augmentation, etc., which are useful to establish a base framework of self-supervised BNNs.
	\item We further propose a guided learning paradigm to boost the performance of self-supervised BNNs. We discuss the roles of contrastive and guided learning in our framework and study the way to use them.
	\item Our proposed framework improves the na\"ive contrastive learning by {\bf 5.5$\sim$15\%} on ImageNet, and we further verify the effectiveness of our learned models on the downstream datasets through transfer learning.
\end{itemize}


\vspace{-0.1in}
\section{Related Work}
\vspace{-0.02in}

\noindent{\textbf{Binary Neural Networks.}} Binary neural networks~\cite{courbariaux2016binarized,rastegari2016xnor,lin2017towards,liu2018bi,phan2020binarizing,martinez2020training,liu2020reactnet} have been widely studied in the recent years. The first work can be traced back to EBP~\cite{soudry2014expectation} and BNNs~\cite{courbariaux2016binarized}. After that, many interesting explorations have emerged. XNOR Net~\cite{rastegari2016xnor} is a representative study that proposed the real-valued scaling factors for multiplying with each of binary weight kernels, this method has become a commonly used binarization strategy in the community and boosted the accuracy of BNNs significantly. Real-to-binary~\cite{martinez2020training} adopted the better training scheme and attention mechanism to propagate binary operation on the activations and obtained better accuracy. ReActNet~\cite{liu2020reactnet} further studied the non-linear activations for BNNs and built a strong baseline upon MobileNet~\cite{howard2017mobilenets}. The proposed method achieved fairly competitive performance on large-scale ImageNet. 

\begin{figure}[t]
	\centering
	    \includegraphics[width=0.48\textwidth]{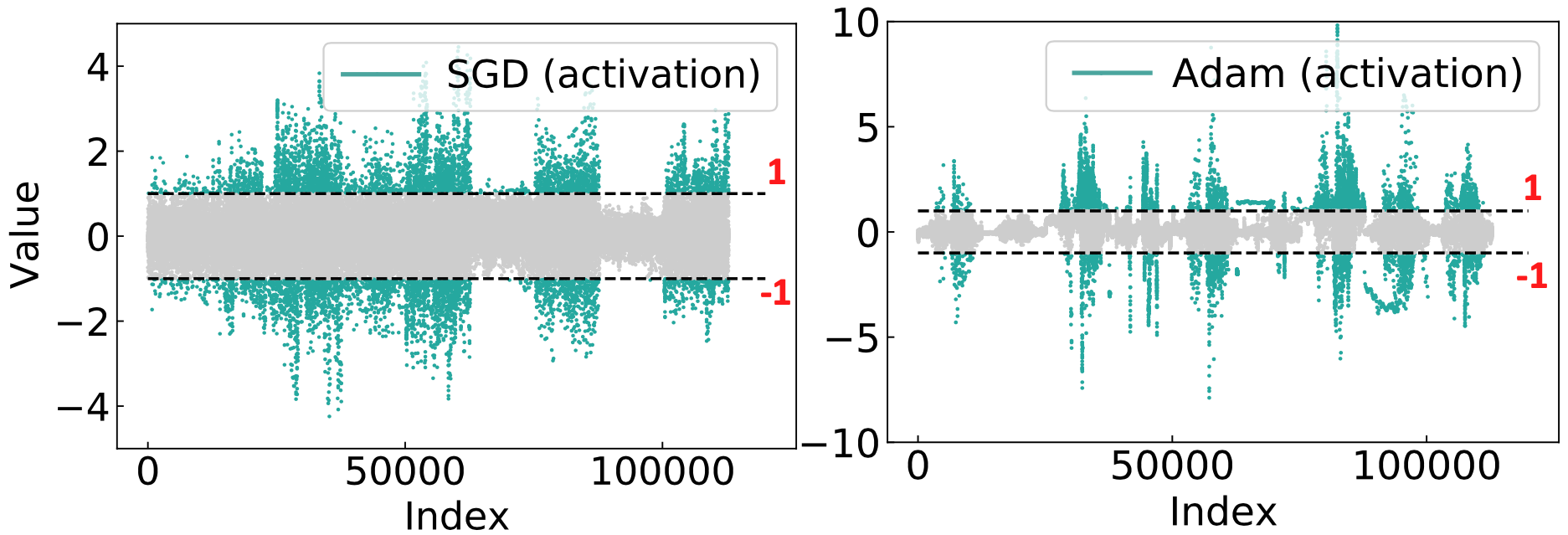}
	   \vspace{-0.15in}
	\caption{Illustration of activation distributions from different optimizers (Left: SGD; Right: Adam). Dotted lines are the up (+1) and low (-1) bounds. We observe that Adam can alleviate activation saturation significantly during training.}
	\label{fig:activate_distribution}
	\vspace{-0.18in}
\end{figure}

\noindent{\textbf{Self-supervised Learning.}} Self-supervised learning (SSL) is a technique that aims to learn the internal distributions and representations automatically through data, meanwhile, without involving any human annotated labels. Early works mainly stemmed from reconstructing input images from a latent representation, such as auto-encoders~\cite{vincent2008extracting}, sparse coding~\cite{olshausen1996emergence}, etc. Following that, more and more studies focused on exploring and designing handcrafted pretext tasks, such as image colorization~\cite{zhang2016colorful}, jigsaw puzzles~\cite{noroozi2016unsupervised}, rotation prediction~\cite{gidaris2018unsupervised}, pretext-invariant representations~\cite{misra2020self}, etc. Recently, contrastive based visual representation learning~\cite{hadsell2006dimensionality} has attracted much attention in the community and achieved breakthroughs and promising results.  Among them, MoCo~\cite{he2020momentum} and SimCLR~\cite{chen2020simple} are two representative methods emerged recently. Also, many interesting works~\cite{oord2018representation,hjelm2018learning,bachman2019learning,tian2019contrastive,shen2020mix,grill2020bootstrap,caron2020unsupervised} have been proposed. In this paper, we expose that the distillation process from a self-supervised strong teacher to the efficient binary student is more effective than learning binary student directly using contrastive learning. A concurrent study SEED~\cite{fang2021seed} also employed self-supervised distillation loss, which can be considered as a contemporaneous work of ours. 

\noindent{\textbf{Self-supervised Learning on BNNs.}} To the best of our knowledge, there are no existing works focusing on exploring BNNs with self-supervised scheme. The proposed approach in this paper has quite appealing advantages on this direction. We will elaborate and validate the proposed method in the following sections. In the network quantization area, Vogel et al.~\cite{8714901} presented a non-retraining method for quantizing networks. This may be the closest work to our study. However, they used the intermediate features of the network based on the valid input samples to supervise the quantization procedure, which is not related to this work, also entirely different from the perspective of our contrastive based or guided learning paradigms.

\section{Optimizer Effects of SSL on BNNs}

\noindent{\textbf{Saturation on Activations and Gradients.}}
We first introduce an activation saturation phenomenon in the self-supervised BNNs scenario. When the absolute values of activations exceed one, the corresponding gradients are suppressed to be zero by the formulation of approximation in the derivative of the {\em sign} function~\cite{ding2019regularizing}. We study this observation for explaining why the optimizer used in self-supervised methods, e.g., MoCo~\cite{he2020momentum} with SGD performs well for real-valued networks, but it is not optimal on binary networks. This exploration can help us determine which optimizer is superior for our proposed self-supervised method. Upon our observation, activation saturation emerges in most layers of a binary network and it always affects the magnitude of gradients critically on different channels. As shown in Fig.~\ref{fig:activate_distribution}, we visualize the activation distributions of the first binary convolution layer of our networks. We can observe that, for the particular input batch images, a large number of activations exceed the bounds of -1 and +1, which causes the gradient passing those neurons to be zero-valued and makes more weights less active.

\begin{figure}[t]
	\centering
	    \includegraphics[width=0.48\textwidth]{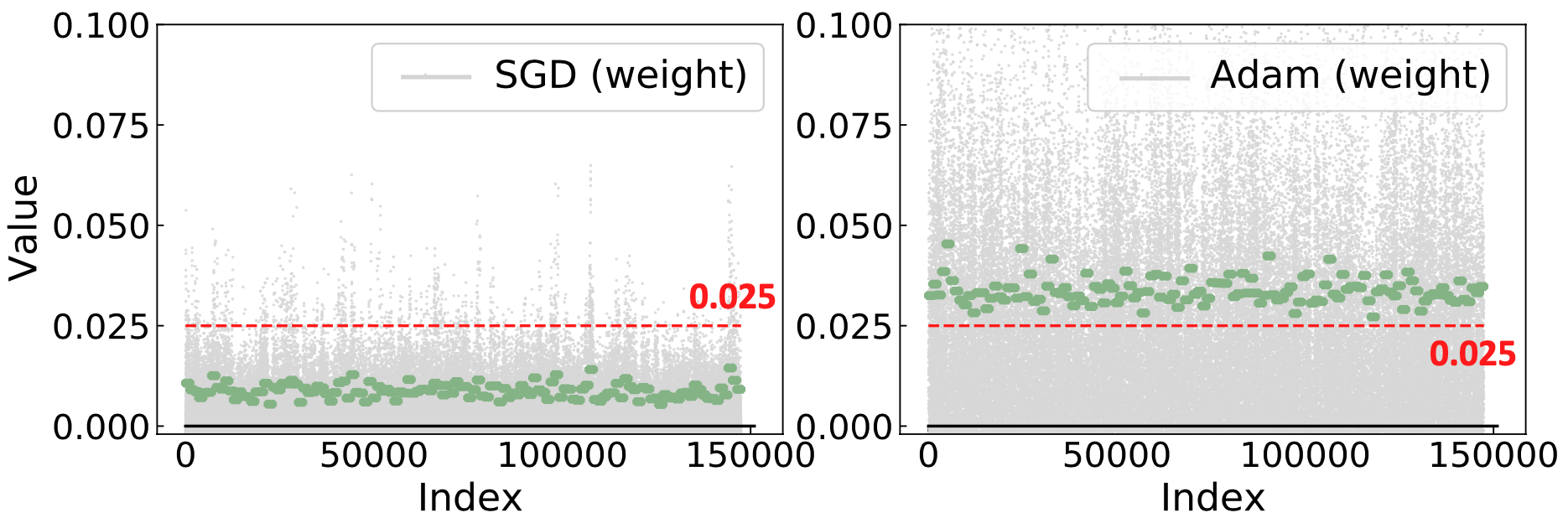}
	    \vspace{-0.19in}
	\caption{Visualization of the weight distributions in the first layer. For clarity, we use \textit{green} hyphens to mark the $l_1$ norm of weights in each channel. The \textit{red} dotted line denotes the reference value (0.025) of weights.}
	\label{fig:weight_distribution}
	\vspace{-0.16in}
\end{figure}

\noindent{\textbf{Different Optimizer Effects.}}
Adam adapts the learning rate according to the historical values and amplifies the gradients with small ones. 
The strength of Adam stems from the regularization effect of second-order momentum, which is crucial to revitalize the inactive weights, i.e., zero-valued ones due to the activation saturation in BNNs as introduced above. Interestingly, Adam can rouse most of the weights to be active again with better generalization ability. The visualization of weight distribution between SGD and Adam in the first layer is shown in Fig.~\ref{fig:weight_distribution}. The red dotted lines are references at the value of 0.025. The green poly-lines are the $l_1$ norm values of weights in each output channel for better comparing to the numerical value between SGD and Adam. It is obvious that Adam contains overwhelmingly larger weights than SGD, which reflects the weights optimized by SGD are not as good as those with Adam.

\begin{figure}[h]
  \centering
  \includegraphics[width=0.3\textwidth]{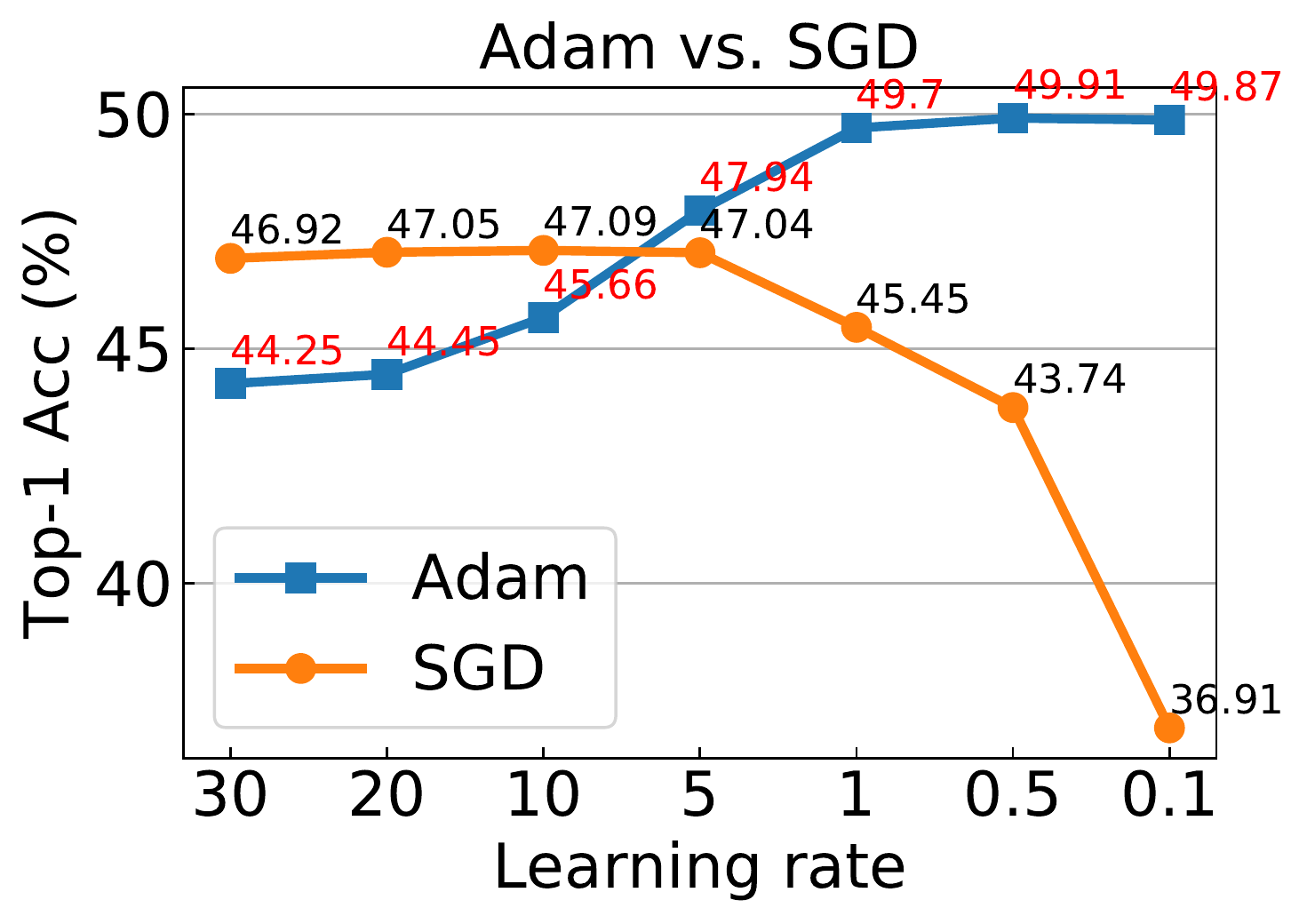}
  \vspace{-0.08in}
  \caption{A comparison of accuracy using SGD and Adam for training backbone networks. The learning rates shown in the figure are in linear evaluation stage and we explore the optimal configurations for them to show the best capability of the two optimizers.}
  \label{fig:improvements}
  \vspace{-0.13in}
\end{figure}

In contrast to the SGD optimizer that only accumulates the first momentum, the adaptive method Adam, predominantly uses the accumulation in the second momentum to amplify the learning rate regarding the gradients with small historical values. SGD with momentum updating is used to help accelerate and dampen oscillations on gradients, it can be formulated as:
$
m_{t}=\beta m_{t-1}+\eta \nabla_{\theta} J(\theta), \ \theta=\theta-m_{t},
$
where $\nabla_{\theta} J(\theta)$ is the gradient and $\beta$ is exponential rate. The updating rule in Adam is defined as:
$
\theta_{t+1}=\theta_{t}-\frac{\eta}{\sqrt{\hat{v}_{t}}+\epsilon} \hat{m}_{t},
$
$\hat m_t$ and $\hat v_t$ denote exponential moving averages of the gradient and the squared gradient, respectively. With $\hat m_t$ drawing $\hat v_t$ of 
the uncentered gradient variance, the update value is normalized to alleviate the discrepancy in the gradient.
Fig.~\ref{fig:improvements} shows the accuracy comparisons on the linear evaluation stage with SGD and Adam-trained backbones in self-supervised learning. It can be observed that, with SGD training, the accuracy decreases when learning rates become smaller. This tendency is consistent with the real-valued model. While with Adam, the accuracy increases dramatically when using smaller learning rates, and the best final accuracy is much higher than the best result from SGD. 

\section{Data Augmentation Adjustments} \label{data_aug}

Our data augmentation strategies mainly inherit from the baseline method MoCo V2~\cite{chen2020improved}. In real-valued networks, heavier augmentations have been proven useful in most cases of contrastive based self-supervised learning. However, considering the limited capability of BNNs to distinguish the same class from different shapes of images, instead of involving more data augmentations, we decrease the transformations' probabilities of {\em ColorJitter} and {\em GaussianBlur} to facilitate the difficulty for BNNs to classify the two images in the same class. Intriguingly, this lite data augmentation strategy can bring an additional $\sim$1.0\% improvement on ImageNet. This reflects that the properties of BNNs are basically different from the real-valued networks, thus the configurations are required to be reconsidered. It also demonstrates the value of this study on exploring self-supervised BNNs. More details are provided in Sec.~\ref{details_exp}. 

\section{Our Approach}

Our roadmap of this paper has three main stages: Firstly, we follow the real-valued self-supervised method with contrastive loss whereas replacing particular configurations to fit the properties of BNNs, such as optimizer, data augmentation, learning rate, etc. These strategies can produce 5.5\% improvement over vanilla MoCo V2 baseline. Then, we propose to adopt an additional guided learning method to enforce representations of BNNs to be similar to the real-valued reference network. This simple strategy can bring an additional $\sim$4\% improvement. Lastly, we remove contrastive loss and solely optimize BNNs with the guided learning paradigm and the performance is further increased by 5.5\%. Several motivations and insights of our proposed method are discussed in the following sections.

\begin{figure*}[t]
  \centering
  \includegraphics[width=1.0\textwidth]{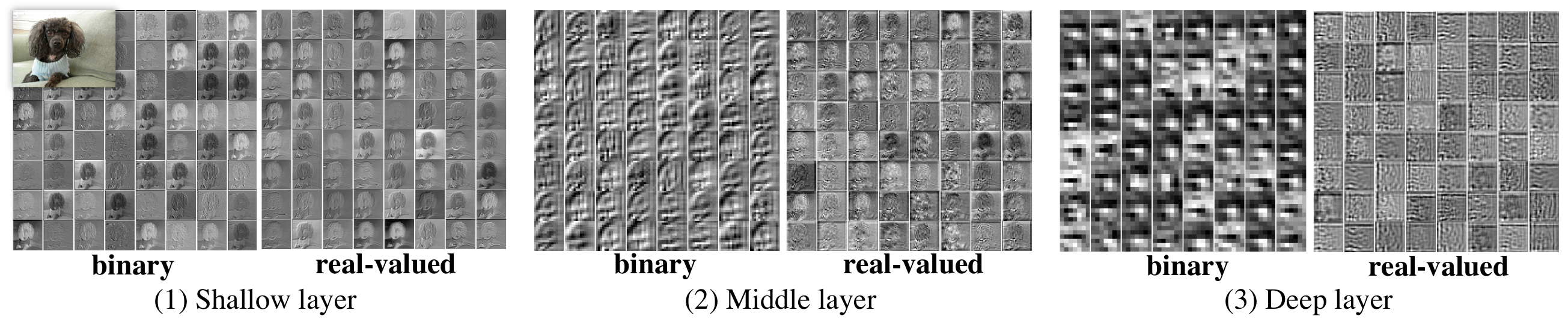}
  \vspace{-0.2in}
  \caption{Illustration of activation distributions on shallow, middle and deep layers of self-supervised binary and real-valued networks. The input image is in the upper left corner of the first subfigure. Interestingly, we observe that at the shallow layer, the semantic representations between these two models are visually similar, while on the middle and deep layers, real-valued representations obviously contain richer information than BNNs in the activation maps, which is beneficial for learning good latent features. This phenomenon motivates us to propose a method for calibrating the high-level distribution of BNNs in a self-supervised learning scenario to arouse its potential.}
  \label{fig:distributions}
  \vspace{-0.15in}
\end{figure*}

\subsection{Preliminaries}

BNNs aim to learn networks that both weights and activations are with discrete values in \{-1, +1\}. In the forward propagation of training, the real-valued activations will be binarized by the sign function:
$\mathcal A_b = {\rm Sign}(\mathcal A_r) = \left\{  
             \begin{array}{lr}  
             - 1 & {\rm if} \ \ \mathcal A_r <0, \\  
             + 1 & {\rm otherwise.}  
             \end{array}  
\right. $
where $\mathcal A_r$ is the real-valued activation of the previous layers calculated from the binary or real-valued convolutional operations. $\mathcal A_b$ is the binarized activation. The real-valued weights in the model will be binarized through:
${\bf W}_b = \frac{||{\bf W}_r||_{l_1}}{n}  {\rm Sign}({\bf W}_r) = \left\{  
             \begin{array}{lr}  
             - \frac{|| {\bf W}_r||_{l_1}}{n}  & {\rm if} \ {\bf W}_r <0, \\ 
             + \frac{||{\bf W}_r||_{l_1}}{n}  & {\rm otherwise.}  
             \end{array} 
\right. $
where ${\bf W}_r$ is the real-valued weights that are maintained as {\em latent} parameters to accumulate the tiny gradients, $n$ is \#weight in each channel.  ${\bf W}_b$ is the weights after binarization. The binary weights will be updated through multiplying the sign of latent real-valued weights and the channel-wise $l_1$ norm ($\frac{1}{n} ||{\bf W}_r||_{l_1}$). The gradient $g$  is calculated with respect to binary weights ${\bf W}_b$: $g_{t} = \nabla_{{\bf W}_{b_t}} J({\bf W}_{b_t}) \cdot 1_{|{\bf W}_{r_t}|<1},$ where $t$ is \#iteration. 

Training BNNs is a challenging task since the gradient for optimizing parameters in the network is approximated and the capacity of models for memorizing all data distributions is also limited.  It is thus worthwhile to discuss that as the {\em sign} function has a bounded range, the approximation to the derivative of the {\em sign} operation will suffer from a vanished gradient issue when the activations exceed the effective gradient range, i.e., $[-1,1]$.

\begin{figure}[t]
  \hspace{0.1in}
  \includegraphics[width=0.46\textwidth]{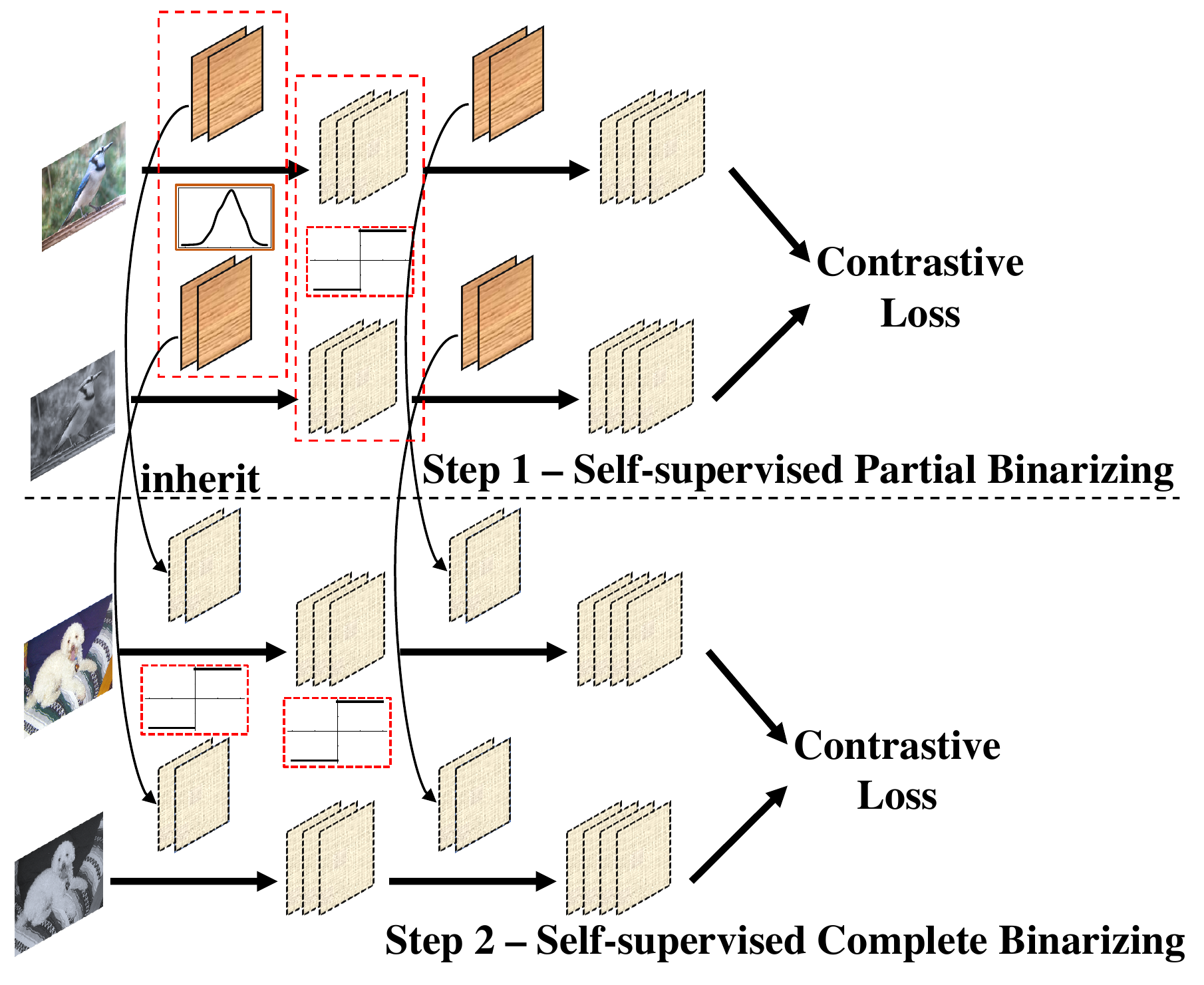}
    \vspace{-0.21in}
  \caption{Illustration of progressive binarization. We take the contrastive based framework as an example, while it is also feasible for the proposed guided learning solely framework.}
  \label{fig:progress_1}
  \vspace{-0.13in}
\end{figure}

\subsection{Real-Value Guided Distillation}

\noindent{\textbf{Self-supervised Contrastive Loss.}}
The conventional contrastive learning uses a standard log-softmax function to apply one positive sample out of $K$ negative samples and it predicts the probability of data distribution as:
\begin{equation}
\begin{gathered}
$$\LL_{CL} =-\log\frac{\exp \left({s\left(\mathbf{v}_{{I}}, \mathbf{v}_{{\hat I}} \right)}/{\tau}\right)}{\exp \left({s\left(\mathbf{v}_I, \mathbf{v}_{\hat I}  \right)}/{\tau}\right)+\sum_{{\hat I}^{\prime}\in {\bf Neg}} \exp \left({s\left(\mathbf{v}_{{I} }, \mathbf{v}_{{\hat I}^{\prime}}\right)}/{\tau}\right)}
$$
\end{gathered} 
\end{equation}
where $\left(\mathbf{v}_{{I}}, \mathbf{v}_{\hat I}\right)$ are two random “views”  of the same image under random data augmentation. $s$ is the cosine similarity or other matching function for measuring the similarity of two representations. $\tau$ is a temperature hyper-parameter.

\noindent{\textbf{Guided with KL-divergence Loss.}} KL-divergence loss is used to measure the degree of how one probability distribution is different from another reference one. We train the BNNs $\text{Bi}_\theta$ by minimizing the KL-divergence between its output ${\bf p}^{{\text {Bi}}_\theta}({x_i})$ and the representation ${\bf  p}^{{\text {Real}}_\theta}({x_i})$ generated by a self-supervised real-valued reference model. The loss function can be formulated as: 
\begin{equation}
	\begin{gathered}
		{\LL_{ KL}}({\text{Bi}_\theta }) =  - \frac{1}{N} {\sum\limits_{i = 1}^N ({{\bf  p}^{{\text{Real}_\theta}}({x_i})/{\tau})\log (\frac{{{\bf p}^{{\text{Bi}_\theta }}({x_i})/{\tau}}}{{{\bf  p}^{{\text{Real}_\theta }}({x_i})/{\tau}}})} }  \hfill \\
	\end{gathered} 
\end{equation}
where $N$ is the number of samples. $\tau$ is a temperature hyper-parameter. Note that the data augmentation strategy should be the same for both binary and real-valued models. In practice, we only optimize with cross-entropy loss as:
\begin{equation}
{{{\LL}}_{CE}}({\text{Bi}_\theta }) =  - \frac{1}{N}{\sum\limits_{i = 1}^N {({\bf p}^{{\text{Real}_\theta }}({x_i})/{\tau})\log } } ({\bf p}^{{\text{Bi}_\theta }}({x_i})/{\tau})
\end{equation}
which is equivalent to $\LL_{KL}$ following MEAL~\cite{shen2019meal,shen2020meal}.

\subsection{Progressive Binarization}

\begin{figure}[t]
  \centering
  \includegraphics[width=0.3\textwidth]{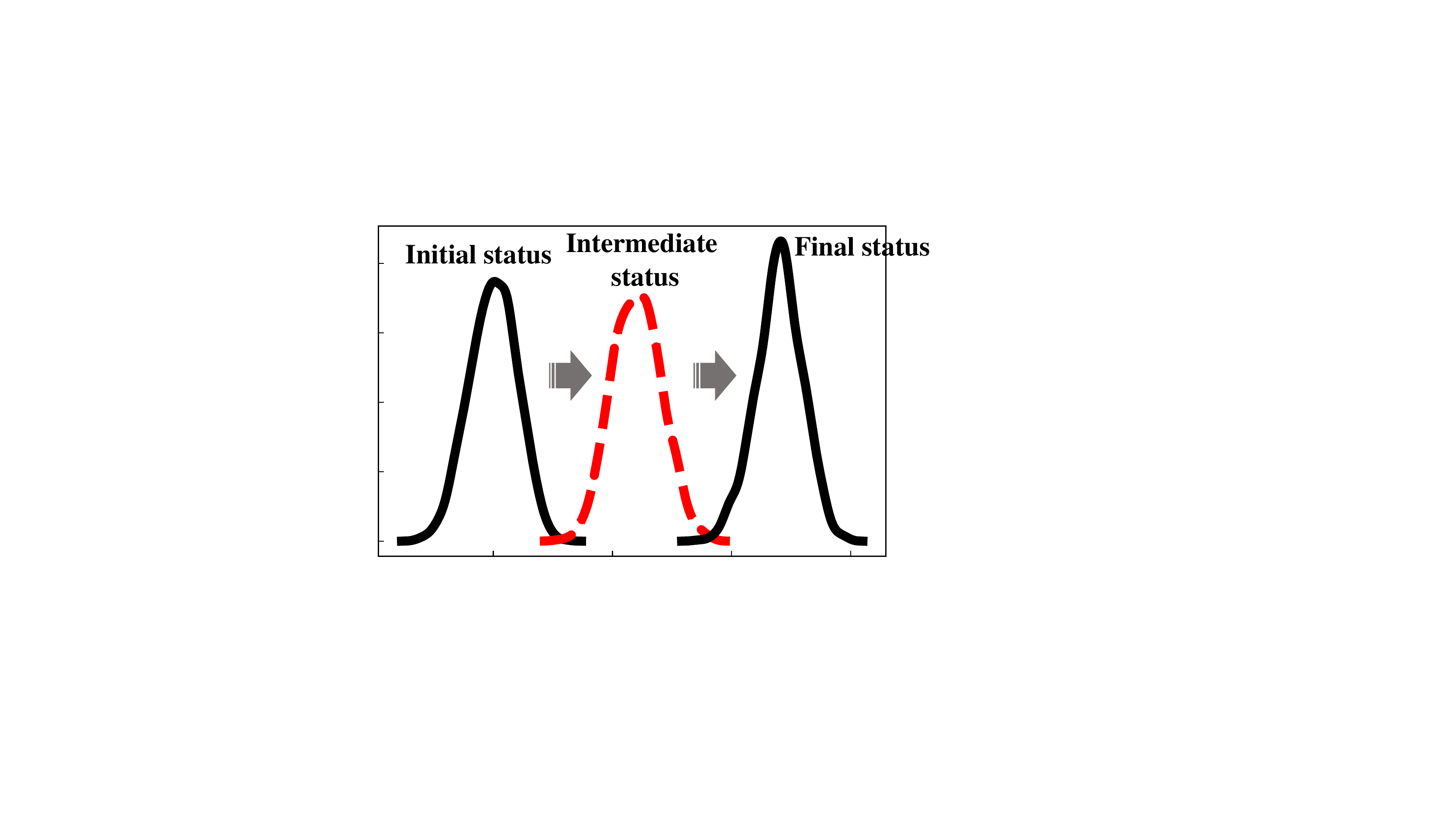}
  \vspace{-0.05in}
  \caption{Illustration of binarization with intermediate status to facilitate self-supervised learning difficulties.}
  \label{fig:progress_vis_new}
  \vspace{-0.15in}
\end{figure}

As illustrated in Fig.~\ref{fig:distributions}, there are many differences on activation distributions between binary and real-valued networks across middle and high-level layers, and real-valued activations always contain more fine-grained details and semantic information of instance and background on representation distributions. As our purpose is to recover the distributions from real-valued networks to binary networks, we propose to adopt a multi-step binarization procedure. The motivation behind this design is straight-forward, as shown in Fig.~\ref{fig:progress_vis_new}, directly recovering distribution from real-valued networks to binary networks is challenging, to facilitate the difficulty of optimization, we first keep partial parameters or weights in the target model to be real-valued, and then binarize them progressively. This strategy is somewhat similar to~\cite{martinez2020training,liu2020reactnet}, while we emphasize that all these previous studies lie in supervised learning, here our objective is a self-supervised contrastive loss or distillation loss. Hence the learning procedure and hyper-parameter design are entirely different from prior works.

\begin{figure*}[t]
  \centering
  \includegraphics[width=0.81\textwidth]{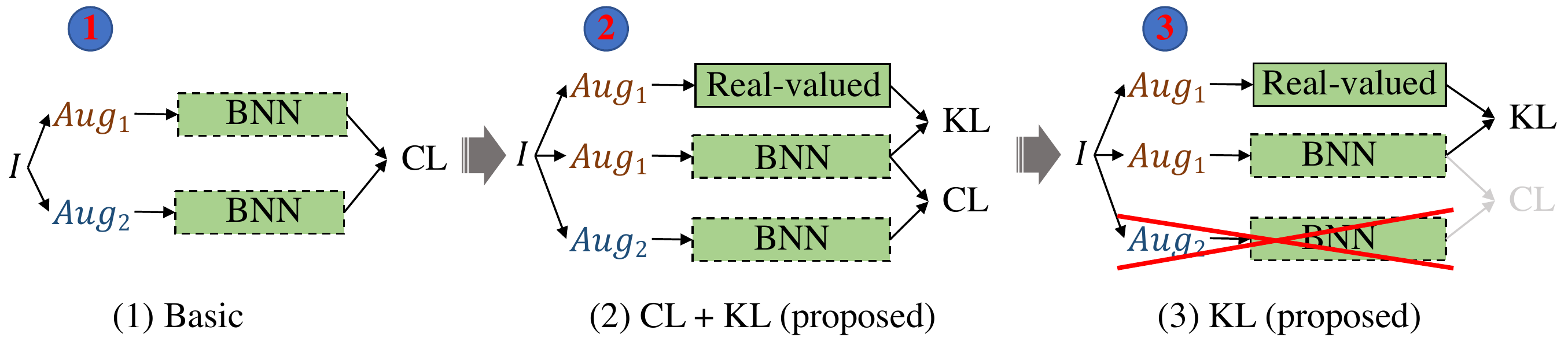}
  \vspace{-0.05in}
  \caption{Illustration of three strategies of our frameworks. ``CL'' denotes the conventional contrastive learning and ``KL'' denotes Kullback–Leibler divergence, i.e. the proposed guided learning. In each subfigure, solid and dashed boxes represent freezing and non-freezing parameters in training, respectively.}
  \label{fig:three_paradigm}
  \vspace{-0.28in}
\end{figure*}

\begin{figure}[t]
  \centering
  \includegraphics[width=0.35\textwidth]{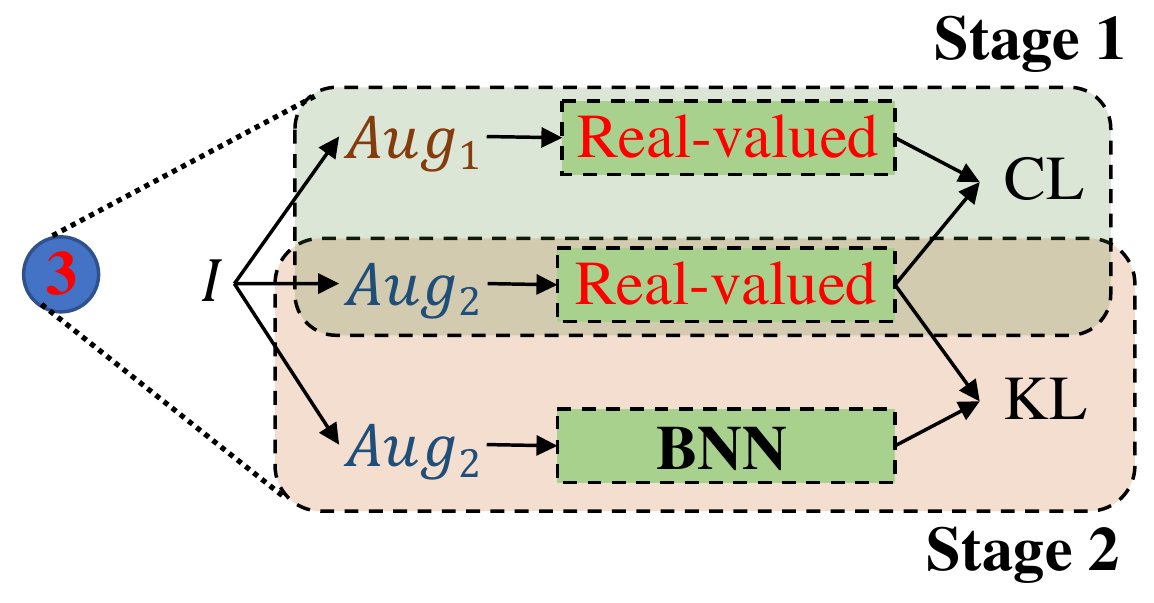}
  \vspace{-0.05in}
  \caption{Illustration of online and offline training of self-supervised real-valued reference network. This is also a more detailed explanation of \ding{174} in Fig.~\ref{fig:three_paradigm}.}
  \label{fig:3_explain}
  \vspace{-0.2in}
\end{figure}

In our method, the initial status is a complete real-valued network, the intermediate status is a partially binarized network with real-valued weights and binary activations, as shown in Fig.~\ref{fig:progress_1}. We train such a network first to obtain the real-valued parameters, then we reuse these pre-trained parameters in the final completely binary models. Since the binarization is modeled by the {\em sign} function during training, hence the binary models can inherit the real-valued parameters as the initialization. This study shows that such a multi-step binarization can facilitate optimization for self-supervised training and obtain significant improvement.

\noindent{\textbf{Weight Decay Strategy.}}  Weight decay is a widely used technique for preventing networks from overfitting. We observe that it is necessary to adopt an appropriate weight decay strategy in different steps of our multi-step training. Since the intermediate status is only a transitional phase, the existing real-valued weights can make model's capacity larger than the final completely-binary status, thus the weight decay is not employed in the first step (or choosing a smaller value of weight decay in this phase). For the second step, weight decay is adopted to avoid overfitting.

\vspace{-0.03in}
\subsection{What Happens If Removing Contrastive Loss}
\vspace{-0.03in}
Since adding guided learning term brings a substantial improvement, we are curious whether guided learning is capable of learning good representations solely. Our observation is surprising, removing contrastive loss gives an additional 5.5\% improvement. We conjecture this is because contrastive and guided learning are basically optimizing with different directions. Guided learning is mimicking the real-valued high-quality representation and recovering the knowledge stored in it, if the reference model is strong enough, the target BNNs can be extremely well-performing, while contrastive loss learns from data itself which explores different patterns (e.g., instance discrimination, colorization, etc.) from guided learning. Therefore, in this work we study the following {\textbf{three schemes}}, as shown in Fig.~\ref{fig:three_paradigm}:

\noindent{\textbf{\ding{172}: Enhanced baseline of contrastive learning.}}

\noindent{\textbf{\ding{173}: Contrastive + guided learning (distillation).}}

\noindent{\textbf{\ding{174}: Guided learning (distillation) only.}}

\noindent{\textbf{Where is the self-supervised real-valued network from?}} There are two ways to obtain the real-valued reference network: (1) online training with the target BNN simultaneously; (2) offline pretraining.  As shown in Fig.~\ref{fig:3_explain}, if we train ``stage 1'' and ``stage 2'' together, this is the online training scheme and the real-valued network will be optimized together with binary network. However, the learning cost will increase significantly since in each individual run we have to include an additional real-valued branch. A simpler and more efficient way is to train ``stage 1'' offline and in advance, then reuse it for all experiments. The offline strategy is utilized in this work of all our experiments. 
Based on the perspective of MEAL V2~\cite{shen2020meal} that better teachers usually distill better students, we choose MoCo V2 800-ep pre-trained real-valued ResNet-50 as our strong teacher model.   

\vspace{-0.1in}
\begin{algorithm}[h]
	\small
	\caption{Online and offline training for self-supervised real-valued reference and target binary networks.}
	\label{inf}
	{\bf Preparation:}
	\begin{algorithmic}[1]
	\State $x_1 = aug_1(x)$ \Comment{$x$ is the input image}
	\State $x_2 = aug_2(x)$ 
	\State ${\bf p}_1^{{\text {Real}}_\theta} = f_{\text {Real}_\theta}(x_1)$   \Comment{$f_{\text{Real}_\theta}$ is the real-valued network}
	\State ${\bf p}_2^{{\text {Real}}_\theta} = f_{\text {Real}_\theta}(x_2)$   
	\State ${\bf p}^{{\text {Bi}}_\theta} = f_{\text {Bi}_\theta}(x_2)$       \Comment{$f_{\text{Bi}_\theta}$ is the binary network}
	\end{algorithmic}
	{\bf  Online Scheme:} 
	\begin{algorithmic}[1]
    \State ${\LL_{}}(f_{\text{Bi}_\theta},f_{\text{Real}_\theta})={\LL_{ CL}}({\bf p}_1^{{\text {Real}}_\theta}, {\bf p}_2^{{\text {Real}}_\theta}) + {\LL_{ KL}}({\bf p}^{{\text {Bi}}_\theta},{\bf p}_2^{{\text {Real}}_\theta})$
    \end{algorithmic}
	{\bf Offline Scheme:} 
	\begin{algorithmic}[1]
    \State Stage 1: ${\LL_{}}(f_{\text{Real}_\theta})={\LL_{ CL}}({\bf p}_1^{{\text {Real}}_\theta}, {\bf p}_2^{{\text {Real}}_\theta})$
    \State ${\bf p}_2^{{\text {Real}}_\theta} = f_{\text {Real}_\theta}(x_2)$ 
    \State ${\bf p}^{{\text {Bi}}_\theta} = f_{\text {Bi}_\theta}(x_2)$ 
    \State $f_{\text{Real}_\theta}$ = $f_{\text{Real}_\theta}$.detach()  \Comment{stop-gradient()}
    \State Stage 2: ${\LL_{}}(f_{\text{Bi}_\theta})={\LL_{ KL}}({\bf p}^{{\text {Bi}}_\theta},{\bf p}_2^{{\text {Real}}_\theta})$
    \end{algorithmic}
\end{algorithm}
    \vspace{-0.15in}

\vspace{-0.1in}
\section{Experiments}
\vspace{-0.03in}

In this section, we first introduce the datasets we used and implementation details for self-supervised pre-training, linear evaluation and transfer learning. Then, we provide extensive ablation studies for each component in our method. Following that, we show our main and transfer results. Lastly, we illustrate activation visualizations to further demonstrate the effectiveness of our proposed method.

\vspace{-0.03in}
\subsection{Datasets and Implementation Details} \label{details_exp}
\vspace{-0.03in}

\noindent{\textbf{Datasets.}} Our experiments are conducted on the widely-used large-scale ImageNet 2012 dataset~\cite{deng2009imagenet}, which contains 1,000 classes with a total number of 1.2 million training images and 50,000 images for validation. For transfer learning, we use PASCAL VOC2007~\cite{everingham2010pascal}, CUB200-2011~\cite{wah2011caltech}, Birdsnap~\cite{berg2014birdsnap} and CIFAR-10/100~\cite{krizhevsky2009learning} benchmarks.

\noindent{\textbf{Data Augmentation.}} As mentioned above, our basic data augmentation follows MoCo V2~\cite{chen2020improved} with no additional operations, but we reduce the probability of {\em ColorJitter} from 0.8 to 0.6 and {\em GaussianBlur} from 0.5 to 0.2. We apply this lite data augmentation strategy to all of our experiments.

\noindent{\textbf{Self-supervised Pre-training.}} We adopt MoCo V2~\cite{chen2020improved} as the baseline self-supervised method. For our distillation solution, we use {\em none} of momentum update, shuffling BN, a memory bank (negative pairs) and contrastive loss. The initial learning rates are 3$\times10^{-2}$ for SGD following~\cite{chen2020improved} and 3$\times10^{-4}$ for Adam, and are reduced with a linear decay through {\em lr} = (initial {\em lr}) $\times$ (1 - epoch / total\_epoch). $\tau$ is set to 0.2 for both contrastive and distillation losses. If no otherwise specified, all networks are trained with 200 epochs.

\noindent{\textbf{Linear Evaluation.}} We freeze all the parameters in the backbone and train a supervised linear classifier using the conventional self-supervised evaluation protocol~\cite{chen2020improved,chen2020simple,shen2020mix}. We train with smaller {\em lr} and 100 epochs, and other hyper-parameters are following the baseline method~\cite{chen2020improved}.

\noindent{\textbf{Transfer Learning.}} We fine-tune the entire network using the weights of our learned models as initializations. We train for 180 epochs with a batch size of 128 and an initial learning rate of 0.01. On PASCAL VOC multi-object classification task, we adopt sigmoid cross-entropy instead of softmax one. We use SGD with a momentum parameter of 0.9 and weight decay of 0.0001. We perform standard random crops with resize and flips as data augmentation during fine-tuning. The training image size is 224$\times$224. At test time, we resize images to 256 pixels and take a 224$\times$224 center crop. When freezing backbone, we solely train the last linear layer as the standard linear evaluation protocol.

\vspace{-0.03in}
\subsection{Ablation Studies}
\vspace{-0.03in}

\noindent{\textbf{Optimizers.}} We study the standard SGD and adaptive optimizer Adam in the pre-training stage. Our results in Fig.~\ref{fig:improvements} shows that Adam can bring about 2.8\% improvement.

\noindent{\textbf{Learning Rate Scheduler.}} Here the learning rate scheduler indicates in the linear evaluation stage. In the training stage, we use a uniform value of 3$\times10^{-4}$ as presented above. The results are shown in Table~\ref{tab:my-table_lr}, we provide results of our three schemes on the {\em lr} range from 30 to 0.05. It can be observed 0.1 is the optimal choice for the Adam optimized models.

\begin{table}[h]
\centering
\caption{One-step results with different {\em lr} in linear evaluation.}
\label{tab:my-table_lr}
\resizebox{0.28\textwidth}{!}{
\begin{tabular}{l|c|c|c}
\hline
{\em lr}  & \ding{172} & \ding{173} & \ding{174} \\ \hline
30   & 44.248  &  47.918 & 54.518  \\
20   & 44.454  & 48.570  &  54.986 \\
10   & 45.662 & 50.054  & 56.050  \\
5  & 47.942  & 51.808  &  57.868 \\
1  & 49.702 &  53.324 &  59.838 \\
0.5  & \bf 49.914  & 53.468  &  59.968  \\
0.1 & 49.870  & \bf 53.484  &  \bf 60.418 \\
0.05& 49.228  &   52.926    &  60.304 \\ \hline     
\end{tabular}
}
\end{table}

\noindent{\textbf{Data Augmentation Effects.}} \label{lite_aug} 
We conducted a comparison using our proposed lite data augmentation and MoCo V2 vanilla strategy, and the same contrastive loss and Adam optimizer are used here. The Top-1/5 results are neatly improved from 49.402/73.152 to 50.410/73.968 on ImageNet.

\begin{table}[t]
\vspace{-0.06in}
\centering
\caption{Comparison of one-step and multi-step binarization.}
\label{tab:progressive}
\resizebox{0.4\textwidth}{!}{
\begin{tabular}{l|c|c|c}
\hline
     & \ding{172} & \ding{173} & \ding{174} \\ \hline
Complete in one step &  49.914 & 53.484  & \bf 60.418  \\            
Multi-step binarization & 52.452  &  56.022 & \bf 61.506 \\\hline
\end{tabular}
}
\vspace{-0.06in}
\end{table}

\setlength\tabcolsep{2pt}
\begin{table}[t]
\caption{Comparison of the Top-1 accuracy on ImageNet with supervised and self-supervised state-of-the-art methods.}
\label{tab:my-table_main}
\resizebox{0.48\textwidth}{!}{%
\begin{tabular}{ccccccc}
\hline
\multicolumn{2}{c}{\multirow{2}{*}{\bf Binary Methods}}        &    \multicolumn{1}{c}{\multirow{2}{*}{\#Epoch}}     & BOPs           & FLOPs          & OPs            & Acc (\%) \\
\multicolumn{2}{c}{}                                       &               & ($\times10^9$) & ($\times10^8$) & ($\times10^8$) & Top-1    \\ \hline
\multicolumn{7}{l}{\bf Supervised Learning:}          \\ \hline                                                                                     
\multicolumn{2}{c}{BNNs~\cite{courbariaux2016binarized}}   & -- & 1.70 &  1.20  & 1.47  &  42.2    \\
\multicolumn{2}{c}{ XNOR-Net~\cite{rastegari2016xnor}}  & -- &  1.70  & 1.41   &  1.67  & 51.2        \\
\multicolumn{2}{c}{MobiNet~\cite{phan2020mobinet}}       &   --   &    --   &    --    &   0.52    &  54.4    \\
\multicolumn{2}{c}{Bi-RealNet-18~\cite{liu2018bi}} & -- &   1.68  &  1.39   &  1.63  &  56.4  \\
\multicolumn{2}{c}{PCNN~\cite{gu2019projection}}   & -- & --  &  -- &1.63 & 57.3 \\
\multicolumn{2}{c}{CI-BCNN~\cite{wang2019learning}}                   &  -- & -- &  -- & 1.63  & 59.9  \\
\multicolumn{2}{c}{Binary MobileNet~\cite{phan2020binarizing}} & -- &   –  &  – &  1.54 & 60.9         \\
\multicolumn{2}{c}{Real-to-Binary~\cite{martinez2020training}} & -- & 1.68 & 1.56   &  1.83   & 65.4    \\
\multicolumn{2}{c}{ MeliusNet29~\cite{bethge2020meliusnet}} & -- &  5.47  & 1.29  & 2.14  & 65.8          \\
\multicolumn{2}{c}{ReActNet~\cite{liu2020reactnet}} &  -- & 4.82           & 0.12           & 0.87           & 69.4     \\ \hline
\multicolumn{7}{l}{\bf Self-Supervised Learning:}                                                                                           \\ \hline
\multicolumn{2}{c}{MoCo V2~\cite{chen2020improved} (baseline)} &  200 & 4.82           & 0.12           & 0.87           & 46.9     \\
\multicolumn{1}{l}{\multirow{3}{*}{\bf Ours}}           &  \ding{172}   &       200        & 4.82           & 0.12           & 0.87           & 52.5     \\
                                      & \ding{173} &       200        & 4.82           & 0.12           & 0.87           &   56.0   \\
                                      & \ding{174} &       200        & 4.82           & 0.12           & 0.87           & \bf 61.5 \\  \hline  
\end{tabular}
}
\vspace{-0.18in}
\end{table}

\noindent{\textbf{Multi-step Binarization.}} Our results are shown in Table~\ref{tab:progressive}, the proposed multi-step strategy generally obtains better accuracy. Whereas, the improvement seems to decrease when the base performance becomes higher, i.e., from \ding{172} to \ding{174}.

\noindent{\textbf{Different Architectures and Strategies.}} 
The results with different backbones are shown in Table~\ref{tab:my-table_overview_1}, we choose XNOR Net~\cite{rastegari2016xnor}, Bi-Real Net~\cite{liu2018bi} and ReActNet~\cite{liu2020reactnet} as our backbones for this ablation study. It can be seen that we obtain substantial improvement across all of these architectures.

\begin{table*}[h]
\centering
\caption{Transfer accuracy on the classification task.}
\label{tab:my-transfer-cls}
	\resizebox{0.68\textwidth}{!}{
\begin{tabular}{lccccc}
 \hline
 & VOC2007 & CUB200-2011 & Birdsnap & CIFAR-10 & CIFAR-100 \\
            \hline
\em From Scratch (Real-valued) & 72.7 &  29.8 & 46.2 &  93.1  & 70.9 \\    
\em From Scratch (Binary) & 50.0 & -- &  -- &  65.9  & 37.2 \\ \hline  
\multicolumn{5}{l}{\em \bf Fine-tune:}                              \\    
MoCo V2 Real-valued (baseline 1)  & 89.6 & 67.3 & 63.6 & 95.3 & 79.3  \\
MoCo V2 Binary (baseline 2)  &  81.0 & 34.4& 34.0 & 89.9  & 69.5  \\
Ours (\ding{172})    & 82.3 & 38.2 & 38.0 &  91.5 & 71.9 \\
Ours (\ding{173})    & 83.5 & 40.5 & 39.2 &  91.3 & 72.3 \\
Ours (\ding{174})    &\bf 86.9 &\bf 50.1 & \bf 45.7 & \bf 92.7 & \bf 74.3 \\ \hline
\multicolumn{5}{l}{\em \bf Freeze backbone:}                            \\     
MoCo V2 Real-valued (baseline 1)  & 86.5 & 51.5  & 22.8 & 86.9  & 60.7  \\
MoCo V2 Binary (baseline 2)  & 79.8 &  23.3 & 20.3 & 79.3 & 56.7  \\
Ours (\ding{172})    & 81.7 & 33.1 & 21.9 &  80.4 & 58.7 \\
Ours (\ding{173})    & 83.1 & 38.4 & 25.6 &  80.7 & 58.8 \\
Ours (\ding{174})    &\bf 86.4 & \bf 47.5 & \bf 34.1  & \bf 82.7 &\bf 61.9\\ \hline
\end{tabular}
}
\vspace{-0.22in}
\end{table*}

\vspace{-0.03in}
\subsection{Main Results}
\vspace{-0.03in}

A summary of our main results is shown in Table~\ref{tab:my-table_main}, we adopt ReActNet as our backbone network. Comparing to the self-supervised baseline MoCo V2, our method outperforms it by 14.6\% with the same training epochs. Promisingly, it can be observed that our results are even comparable to some recently proposed supervised methods, such as Bi-RealNet-18~\cite{liu2018bi}, CI-BCNN~\cite{wang2019learning} while only containing about 1/2 OPs to them. The results demonstrate the great potential of our self-supervised BNN method on real-world applications where annotation and memory are both scarce.

\begin{figure}[t]
  \centering
  \includegraphics[width=0.44\textwidth]{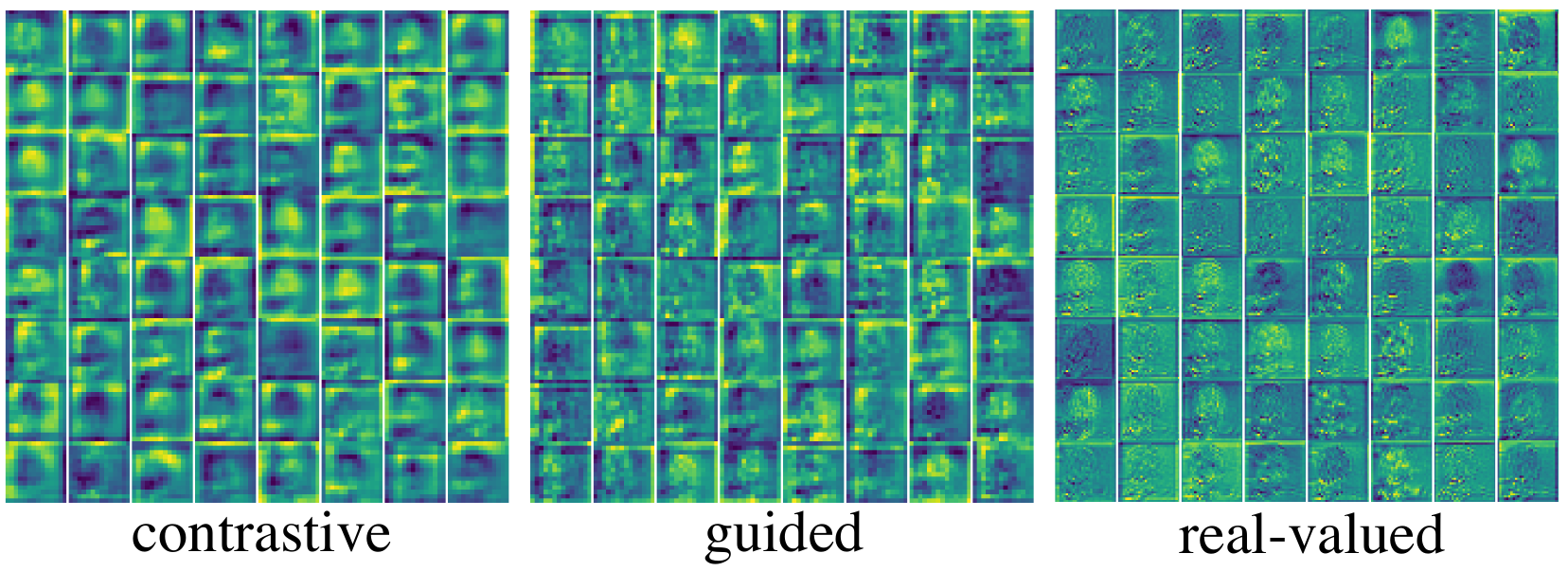}
  \vspace{-0.08in}
  \caption{Illustration of contrastive and guided learned activation maps from the same layer of models. Visually, it can be identified that the quality of activation maps from contrastive learning to guided learning is improved with more details.} 
  \label{fig:activation_results}
    \vspace{-0.18in}
\end{figure}

\noindent{\textbf{Visualization.}}
To better understand where the boost comes from in our distillation method, we further visualize the activation maps of contrastive and guided learned models at the same level of layers. As shown in Fig.~\ref{fig:activation_results}, in each group, we visualize the first 64 channels in those layers. Visually, it can be recognized that the quality of activation maps from contrastive learning to guided learning is improved significantly with more details, and guided learning results are more close to the real-valued ones.

\vspace{-0.03in}
\noindent{\textbf{More Training Epochs.}} In real-valued scenario of self-supervised learning, more training budget always obtains a significant improvement. For example, SwAV~\cite{caron2020unsupervised} achieves 0.7\% gain when training from 200 to 400 epochs. However, we also train our model with 400 epochs but we found the improvement is marginal (from 61.5\% to 61.8\%). We conjecture the reason is that our distillation based framework utilizes neither positive nor negative pairs, the binary student basically recovers the teacher's capability, so it is bounded by teacher's ability rather than the training budget.

\vspace{-0.03in}
\noindent{\textbf{Training Cost Analysis.}} Compared to the self-supervised baseline method, our main extra training cost is the learning procedure of generating the self-supervised real-valued model. As we adopt the offline strategy in our framework, we only need to train it once, hence if not considering this pre-training process, our total computational cost is nearly the same as the baseline MoCo V2. 

\noindent{\textbf{Why Solely Using Distillation Loss is Better Than Combining with Contrastive Loss for Self-supervised BNNs.}
Intuitively, distillation loss forces BNNs to {\em mimic the reference network’s predictive probability}, while contrastive learning tends to {\em discover the latent patterns from the data itself}. Our Fig.~\ref{fig:activation_results} evidences that in binary scenario, contrastive loss is relatively weaker than distillation loss to learn fine-grained representations, and the semantics are also vague. Combining both of them may not be an optimal solution due to the discrepancy of the optimization spaces.

\noindent{\textbf{Transfer Learning.}} 
It is critical to further verify the transferability of our learned parameters from different learning schemes. We follow the conventional self-supervised fine-tuning evaluation protocol for this study. The summary of our transfer results is provided in Table~\ref{tab:my-transfer-cls}, all the network structures in this table are MobileNet-like ReActNet. Our results of \ding{172} can be regarded as the stronger contrastive baseline. ``From scratch'' denotes we train networks with the randomly initialized parameters and we show them here for the reference purpose. Generally, our transfer results are consistent with their linear evaluation performance on ImageNet. In particular, the improvement from \ding{173} to \ding{174} is dramatically higher than that from \ding{172} to \ding{173} across different datasets. Moreover, we observe our best result is even close to the self-supervised real-valued baseline.

\vspace{-0.08in}
\section{Conclusion}
\vspace{-0.05in}
It is worthwhile considering how to train a robust and accurate self-supervised binary network. In this work, we have summarized and explained several behaviors observed while training such networks without labels. We focused on how optimizer, learning rate scheduler and data augmentation encourage representations and affect the performance in building a base BNN framework. We further proposed a guided learning paradigm enabled by a real-valued reference network to distill the target binary network training, and exposed such learning strategy can obtain better results comparing to both the contrastive learning and even supervised learning BNNs scheme. We attribute the proposed superior training scheme to its ability of mimicking the high quality of the reference network's representation. Finally, we performed extensive ablation experiments on each component of our method. 
Moreover, our trained parameters can be crucial for many downstream tasks that depend on a good representation, such as fine-grained recognition, etc.

{\small
\bibliographystyle{ieee_fullname}
\bibliography{egbib}
}

\end{document}